\documentclass[letterpaper, 10 pt, conference]{ieeeconf}  

\IEEEoverridecommandlockouts                              

\title{\LARGE \bf
ShapeGrasp: Zero-Shot Task-Oriented Grasping \\ with Large Language Models through Geometric Decomposition
}

\author{Samuel Li, Sarthak Bhagat, Joseph Campbell, Yaqi Xie, Woojun Kim, Katia Sycara, Simon Stepputtis
\thanks{All authors are with the Robotics Institute, Carnegie Mellon University.
        {\tt\small \{swli, sarthakb, jacampbe, yaqix, woojunk, sycara, sstepput\}@andrew.cmu.edu}}%
}

\usepackage{times}

\usepackage{multicol}
\usepackage[bookmarks=true]{hyperref}
\usepackage{graphicx}
\usepackage{subcaption}
\usepackage{booktabs}
\usepackage{pifont}
\usepackage{textcomp}
\usepackage[table, dvipsnames]{xcolor}
\usepackage{tabularx}
\usepackage{siunitx}
\usepackage{amsmath}
\usepackage{amssymb}

\usepackage{array,etoolbox}
\preto\tabular{\setcounter{magicrownumbers}{0}}
\newcounter{magicrownumbers}
\newcommand\rownumber{\stepcounter{magicrownumbers}\arabic{magicrownumbers}}

\usepackage{amsmath,amsfonts,bm}

\def\eqref#1{equation~\ref{#1}}

\def\1{\bm{1}}

\def\vg{{\bm{g}}}

\def\mI{{\bm{I}}}

\def\mM{{\bm{M}}}

\DeclareMathAlphabet{\mathsfit}{\encodingdefault}{\sfdefault}{m}{sl}
\SetMathAlphabet{\mathsfit}{bold}{\encodingdefault}{\sfdefault}{bx}{n}

\def\gC{{\mathcal{C}}}

\def\gG{{\mathcal{G}}}

\def\gP{{\mathcal{P}}}

\def\sR{{\mathbb{R}}}

\newcommand{\cmark}{\ding{51}} 

\begin{document}

\maketitle
\thispagestyle{empty}
\pagestyle{empty}

\maketitle

\begin{abstract}
Task-oriented grasping of unfamiliar objects is a necessary skill for robots in dynamic in-home environments.
Inspired by the human capability to grasp such objects through intuition about their shape and structure, we present a novel zero-shot task-oriented grasping method leveraging a geometric decomposition of the target object into simple, convex shapes that we represent in a graph structure, including geometric attributes and spatial relationships. 
Our approach employs minimal essential information -- the object's name and the intended task -- to facilitate zero-shot task-oriented grasping. 
We utilize the commonsense reasoning capabilities of large language models to dynamically assign semantic meaning to each decomposed part and subsequently reason over the utility of each part for the intended task.
Through extensive experiments on a real-world robotics platform, we demonstrate that our grasping approach's decomposition and reasoning pipeline is capable of selecting the correct part in $\mathbf{92\%}$ of the cases and successfully grasping the object in $\mathbf{82\%}$ of the tasks we evaluate.
Additional videos, experiments, code, and data are available on our project website: \href{https://shapegrasp.github.io/}{\texttt{https://shapegrasp.github.io/}}.

\end{abstract}

\IEEEpeerreviewmaketitle

\section{Introduction}

\begin{figure*}
    \centering
    \includegraphics[width=1\linewidth]{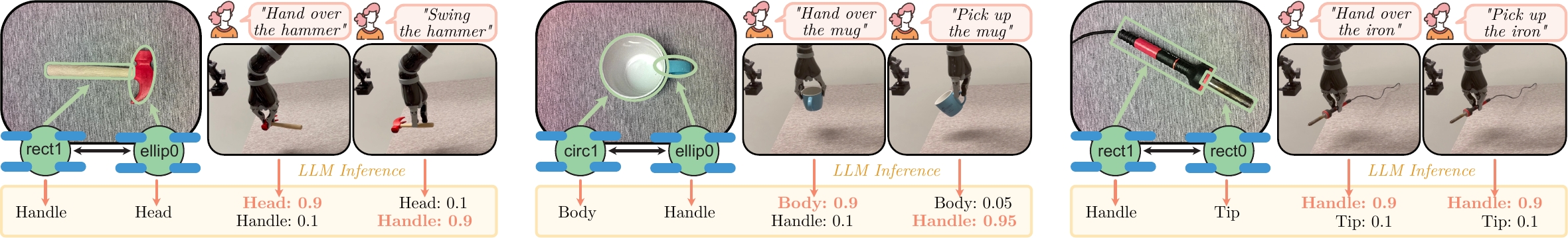}
    \caption{Different resulting grasps given our shape-based inference pipeline. Parts in orange-boldface are ultimately grasped. Green circles and blue lines represent the part-graph decomposition with each entity's associated attributes.}
    \label{fig:tog}
\end{figure*}

In-home environments present a significant challenge for real-world robotics, primarily due to their highly unstructured nature.
As a result, these environments frequently contain novel objects not present in the robot's training environment; however, interacting with such objects, particularly grasping them in a task-dependent manner, is a necessary skill.
Grasping an object in a way that facilitates a certain task, namely \textit{task-oriented grasping}, requires a system to not only detect an object but also to reason over the utility of its parts. 
For example, when picking up a hammer with the goal to ``hand it over'' (see Fig.~\ref{fig:tog}, left), the robot should grasp the hammer by the head to promote ease and safety for the human receiving it.
Large Language Models (LLMs) provide the capability of such commonsense reasoning and can be utilized for task-oriented grasping with only a minimal set of contextual information, namely the object's name and desired task~\cite{murali2020taskgrasp, lerftogo2023}.
However, zero-shot task-oriented grasping remains challenging, particularly since current approaches are computationally expensive and may require additional information beyond the object's name and intended task for high performance, such as desired object part names~\cite{murali2020taskgrasp, lerftogo2023}, thus limiting their zero-shot performance. 




In this work, we propose \texttt{ShapeGrasp}, \textbf{a robust grasping framework based on a geometric decomposition of the target object and shape-based semantic part reasoning leveraging LLMs, capable of conducting task-oriented reasoning and grasping of novel objects}.
This approach is inspired by the human ability to interact with a novel object by analyzing its geometric composition, relating it to prior knowledge, and inferring each part's utility~\cite{OpdeBeeck10111} before utilizing this knowledge to identify a suitable part for an intended task.
Unlike prior approaches, we introduce a multi-step reasoning approach that leverages a symbolic graph of basic shapes and geometric attributes that compose the object and a multi-stage LLM prompt that a) assigns semantic meaning to each part of the object given its name and b) reasons over the affordances and task utility of each part to select the most suitable part for the specific task.
\begin{figure}
    \centering
    \includegraphics[width=1\linewidth]{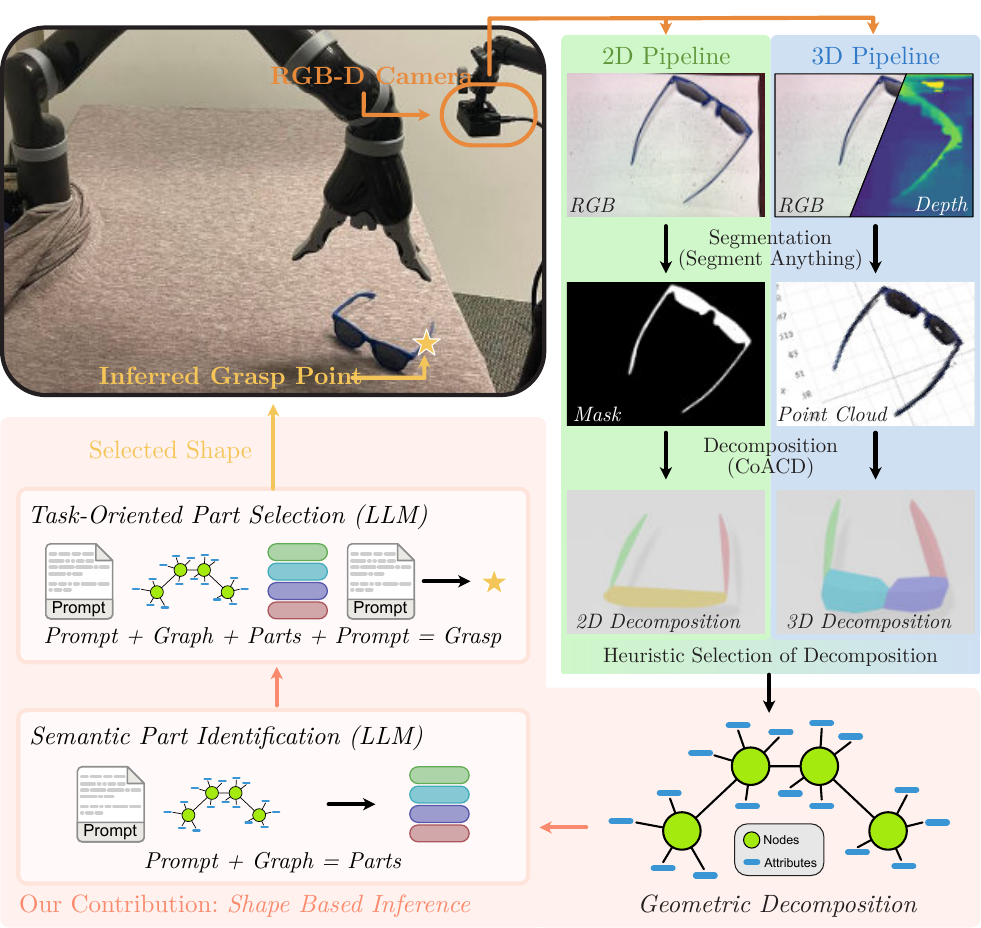}
    \caption{\textbf{The \texttt{ShapeGrasp} Pipeline}: Given a target object, our RGB+D-based approach decomposes the object into basic convex parts. We propose a heuristic approach to decide which decomposition to use before converting it into a shape graph, allowing an LLM to utilize its commonsense reasoning to identify part semantics and task suitability.
    }
    \label{fig:overview}
\end{figure}
Figure~\ref{fig:overview} provides a high-level overview of the functionality proposed in \texttt{ShapeGrasp}. 
Starting with a passive monocular RGB+D image, we segment the target object using a pre-trained Segment Anything Model (SAM)~\cite{kirillov2023segany}, receiving a 2D image mask and the object's relevant point cloud by applying the mask to the depth image. 
We then calculate two approximate convex decompositions using a pre-trained CoACD~\cite{Wei_2022}: 1) a 3D decomposition using the masked point cloud and 2) a 2D decomposition using the image's SAM mask.
\texttt{ShapeGrasp} heuristically identifies a suitable object decomposition from the two decomposition procedures, approximates each convex hull with a basic geometric shape, including a combination of rectangles, circles, triangles, and ellipses, and represents these parts and their spatial relationships as a graph.
We then utilize an LLM to determine the semantic meaning of each shape given the name of the object and identify the best part to grasp using the assigned fine-grained semantic part labels and their suitability for the desired task.
Through this process, inspired by the Chain of Thought~\cite{wei2022chain} approach, we improve the reasoning capabilities of our method. 
Finally, the object is grasped at the identified part by calculating the centroid and principle components of its masked point cloud to determine the grasping location and orientation. 
Our work provides the following contributions:
\begin{itemize}
    \item We introduce \texttt{ShapeGrasp} -- a novel approach for zero-shot task-oriented grasping leveraging a previously unseen object's geometry by creating a graph of basic geometric shapes from decomposed convex parts, allowing for improved reasoning over each part's semantic significance and task-oriented utility.
    \item Our proposed method is a lightweight approach in both the vision-based graph construction stage and reasoning stage compared to other zero-shot grasping methods, utilizing only basic, mininal information and a single static RGB+D image. 
    \item Through extensive experiments on a real-world robotic platform, we demonstrate that our proposed approach outperforms current state-of-the-art methods methods.
\end{itemize}

\section{Related Works}





Robotic grasping \cite{Zhang2022RoboticGF} aims to determine the best way to grasp objects and conventionally includes two approaches: the analytical approach~\cite{ten2017grasp, zapata2019fast} and the data-driven approach~\cite{jang2017end,  zhao2021regnet, alliegro2022end}. 
The \textbf{analytical approach} aims to identify an appropriate grasp pose through analytic models that consider geometric conditions.
For instance, \cite{zapata2019fast} analyzes the geometry of the point cloud and identifies the appropriate grasping points based on a set of geometric conditions. 
Prior works have shown that understanding geometric information dramatically benefits 
robotic grasping~\cite{ten2017grasp, zapata2019fast}, as well as tasks including 3D geometry reasoning~\cite{paschalidou2020learning} and building synthetic tools~\cite{liu2018physical}; however, these approaches require the availability of 3D geometry, which is computationally expensive and sensitive to noise, thereby restricting their applicability in real-world environments. 
On the other hand, the \textbf{data-driven approach}~\cite{jang2017end, zhao2021regnet, alliegro2022end, Agarwal2023DexterousFG} train models to predict grasp points. Prior works propose end-to-end grasp detection networks for partial, noisy point clouds \cite{zhao2021regnet, alliegro2022end} and for leveraging semantic information \cite{jang2017end} in a supervised manner. In addition, techniques based on leveraging a semantic knowledge graph \cite{murali2021same, pmlr-v232-bhagat23a, Bhagat2023KnowledgeGuidedSA}, shape segmentation techniques \cite{lin2020using}, and a physics simulator~\cite{Fang2018LearningTG} have been considered. 
These approaches are limited in their ability to generalize to unseen objects due to requiring computationally expensive training.


Besides the approaches mentioned above, LLMs and VLMs (Vision-Language Models) have recently been successfully utilized in robotics grasping tasks due to their reasoning ability~\cite{mirjalili2023lan, lerftogo2023, tang2023graspgpt, tang2023task}. 
Closest to our work is LERF-TOGO~\cite{lerftogo2023}, which utilizes a vision-language model in a zero-shot fashion to output a ranking of sampled grasping points given a natural language query indicating the object as well as the part-name that should be grasped.
Specifically, LERF-TOGO reconstructs a 3D object mask based on DINO embeddings  \cite{caron2021emerging} and then uses it to output a ranking of grasping points via language models with conditional LERF \cite{kerr2023lerf}. 
While LERF-TOGO requires detailed natural language queries containing the object and object part name, as well as many views to reconstruct the scene and 3D mask, our approach, \texttt{ShapeGrasp}, is able to generate grasping points with minimal semantic information about the object, based solely on a single passive RGB+D top-down view of the scene. 
\texttt{ShapeGrasp} achieves this by decomposing the object into a graph containing geometric shapes, their spatial relationships, and basic attributes, before utilizing the graph as a prompt component for an LLM. 
Thus, our approach leverages the advantages of the analytic approach—\textit{understanding geometric information} \cite{Ding2023KnowledgeCG}---and the large models-based approach---\textit{the zero-shot reasoning ability}  \cite{Huang2022TowardsRI}---which has been shown to be effective.
LLMs and VLMs have successfully been applied to the robotics domain~\cite{ahn2022can, liu2024ok, huang2023inner, stepputtis2020language, zhou2023learning}, including
task-oriented grasping~\cite{lerftogo2023, mousavian20196}. 
In such context, LLMs have been integrated into planning procedures in various ways, such as providing semantic knowledge \cite{ahn2022can} and performing complex reasoning in the form of an inner monologue \cite{huang2023inner}. 
In addition, \cite{liu2024ok} leverages the zero-shot reasoning capabilities of VLMs for the pick-and-drop task requiring object detection, navigation, and grasping. 
Despite the benefits of utilizing LLMs and VLMs, which include no need for additional training and providing commonsense reasoning capability, naive utilization of LLMs and VLMs still have limitations stemming from their inherent shortcomings, such as indecisiveness, lack of domain knowledge, hallucination, and the black-box problem~\cite{Pan2023UnifyingLL}. 
To address these limitations and enhance reasoning, we infuse LLMs with structured knowledge in the form of a symbolic graph that captures an object's geometric composition by describing, among other properties, the shape and size of decomposed convex parts, and the spatial relationships between them. 
This infusion has been shown to be effective in preventing LLMs from deviating into the realm of fictitious information, thereby ensuring a connection to factual data \cite{Ding2023KnowledgeCG, moiseev-etal-2022-skill, Pan2023UnifyingLL, Yang2023ChatGPTIN} and enhancing reasoning capabilities for task-oriented grasping. 




\section{Shape-Based Grasping}
In this section, we introduce \texttt{ShapeGrasp}, our approach to zero-shot task-oriented grasping of novel objects by leveraging a graph of basic geometric shapes that compose the object.
Given an RGB+D input image $\mI \in \sR^{H \times W \times C}$, our approach $\vg, \theta = f_{SG}(\mI)$ estimates a 
grasp location $\vg \in \sR^3$ and rotation $\theta \in \sR^1$ for the robot to pick up the object. 

The function $f_{SG}(\dots)$ represents our modular grasping pipeline, \texttt{ShapeGrasp}, composed of the following modules:
First, we introduce our approach to segmenting and retrieving convex decompositions of the target object (see Sec.~\ref{sec:segmenting}) by using a pipeline of pre-trained models. 
Then, we discuss our first contribution, selecting a suitable decomposition through an automatic heuristic (see Sec.~\ref{sec:extracting_shapes}). 
Finally, we introduce our novel approach of reasoning over the geometric composition of the target object by using a graph-based representation of the object's decomposed parts and utilizing an LLM for a multi-stage reasoning process (see Sec.~\ref{sec:grasp_selection}).

\subsection{Segmenting the Object}
\label{sec:segmenting}
In this section, we introduce our image-processing pipeline utilizing SAM~\cite{kirillov2023segany} to retrieve an object mask and CoACD~\cite{Wei_2022} to generate a convex decomposition of the object.

\subsubsection{Retrieving the Object Mask and Point Cloud}

To obtain the segmentation mask for input image $\mI$, we utilize user input defining a set of \textit{``in-points''} on the target object, used by \textit{Segment Anything Model} (SAM) to retrieve the object's full 2D segmentation mask $\mM$. 
Following this step, in-points are discarded, and a point cloud $\gP$ of the object is retrieved utilizing the mask and depth information in our input image $\mI$.
We posit that the particular method for retrieving such object masks is not central to our overall approach and can be facilitated by various alternative approaches.

\begin{figure}
    \centering
    \includegraphics[width=1\linewidth]{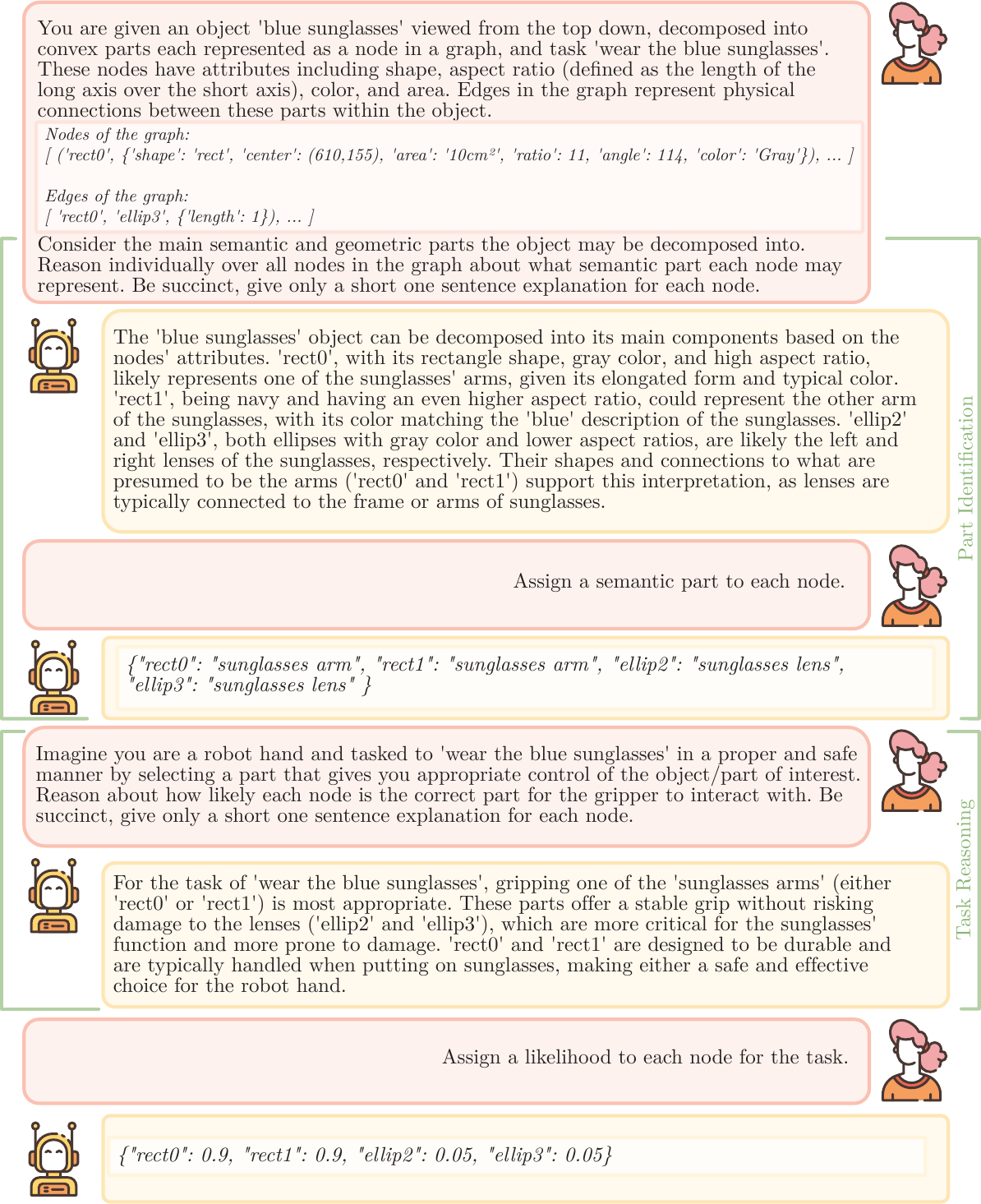}
    \caption{\texttt{ShapeGrasp} Prompting: Given a geometric decomposition graph, we infer a suitable grasping point through a chain of four consecutive prompts: two for semantic part identification, and two for task-oriented reasoning and selection. Ablations of these prompts can be found in Table~\ref{tab:ablation_study}. }
    \label{tab:prompt}
\end{figure}

\subsubsection{Convex Decomposition with CoACD}
Given an object mask $\mM$ and point-cloud $\gP$, we independently retrieve a convex decomposition for each of the two inputs.
To this end, we utilize CoACD, which is a recent approach for the convex decomposition of 3D meshes, specifically designed to retain fine-grained object features, which are important to preserving the original functionality, particularly in interactive settings. 
To utilize this approach, we convert the 2D mask $\mM$ into a 3D mesh by interpreting it as a plane and retrieving a mesh from point cloud $\gP$ through voxelization.  
The depth data utilized in the 3D pipeline allows for more intricate decomposition with CoACD, uncovering features at various elevations within the object that might be overlooked by the 2D method.
However, 2D decompositions are robust to depth inaccuracies and provide a fast approximation that can be beneficial, particularly for concave objects like mugs. 
We compute two separate convex decompositions, one from the 2D mask and one from the 3D point cloud, termed $\gC_{2D}$ and $\gC_{3D}$, respectively, as each decomposition exhibits a set of different desirable properties. 
Selecting the appropriate decomposition depends on the object the system interacts with.
In the next section, Sec.~\ref{sec:extracting_shapes}, we discuss our automated heuristic utilized for this purpose.

\subsection{Geometric Decomposition}
\label{sec:extracting_shapes}

In this section, we discuss the following contributions: a) our heuristic approach to finding a suitable decomposition of a particular object, and b) how we create object-graph $\gG$ that relates part shapes with each other and stores their attributes.

\subsubsection{Heuristic Selection of Decomposition}
To address the problem that certain objects (e.g., concave mugs) are inherently unsuitable for convex decomposition, while other objects may have little visual features or reflective surfaces that result in low-confidence depth maps,
we introduce a heuristic $\gC^* = h(\gC_{2D}, \gC_{3D})$ to choose the optimal decomposition $\gC^*$ from the two described decomposition procedures.
We first lower the decomposition threshold from initial value $\gamma$ until both the 2D and 3D pipelines result in more than a single part. After that, our heuristic $h(\dots)$ chooses a preferred decomposition given the following criteria: If the 3D decomposition $\gC_{3D}$ results in more than a set threshold of parts, $\omega$, 
or if
the percentage of depth points with high confidence $\alpha$ is too low, 
our heuristic $h(\dots)$ chooses the 2D decomposition $\gC_{2D}$.
Formally, $h(\dots)$ selects between 2D and 3D decompositions at valid thresholds as follows:
\begin{equation}
    h(\gC_{2D}, \gC_{3D})= 
\begin{cases}
    \gC_{3D}, & \text{if } |\gC_{3D}|\leq \omega ~ \land ~ \text{conf.} \geq \alpha \\
    \gC_{2D}, & \text{otherwise}
\end{cases}
\end{equation}
where $|\gC|$ is the number of parts found in the respective decomposition. 
Section~\ref{sec:threshold} provides empirical evidence for our heuristic parameters.
With the selected decomposition $\gC^*$, we create the object graph $\gG$ that describes the object's decomposed parts and their spatial relationships.
 
\subsubsection{Structured Object-Graph Creation}
After selecting an appropriate decomposition $\gC^*$, we project it back onto the original input image $\mI$ and create an object-graph $\gG$ describing the composition of the target object.
Each decomposed part is represented as a node in the graph accompanied by attributes derived from the segmented image. 
The primary attribute is an approximating shape primitive, chosen from an isosceles triangle, rectangle, circle, or ellipse.
To select an appropriate shape primitive, we approximate each convex hull with a simplified polygon given a pre-defined approximation threshold $\varepsilon$. 
The resulting points from this simplification dictate the fitted shape as follows:
\begin{itemize}
    \item \textbf{Isosceles triangles} are formed by modifying any three points to equalize the leg lengths and adjust the base. 
    \item \textbf{Rectangles} are formed by finding the rotated rectangle of the minimum area enclosing part.
    \item \textbf{Circles} are identified by measuring the shape factor, or the ratio of the part's area to the area of the bounding circle, with a threshold of $0.9$ for circle classification. 
    \item \textbf{Ellipses} are formed by fitting the part inside a rectangle (see above) if the fit reduces errors further.
\end{itemize}
These geometric shapes allow for the determination and inclusion of additional attributes, namely the aspect ratio, calculated based on the long and short sides of each shape (or major/minor axes), and the angle of the long side. The original decomposed convex hulls provides the centroid and area attributes.
Object color is also incorporated an as attribute, and is derived by bucketing the RGB color spectrum based on the standard $16$ web colors and selecting the most prevalent color. 
Edges within graph $\gG$ are drawn to connect nodes whose convex parts share boundaries or intersect in the segmentation. The length of the connection is included as the edge attribute. 

\subsection{Grasp Inference through Shape Reasoning}
\label{sec:grasp_selection}

\begin{figure}
    \centering
    \includegraphics[width=1\linewidth]{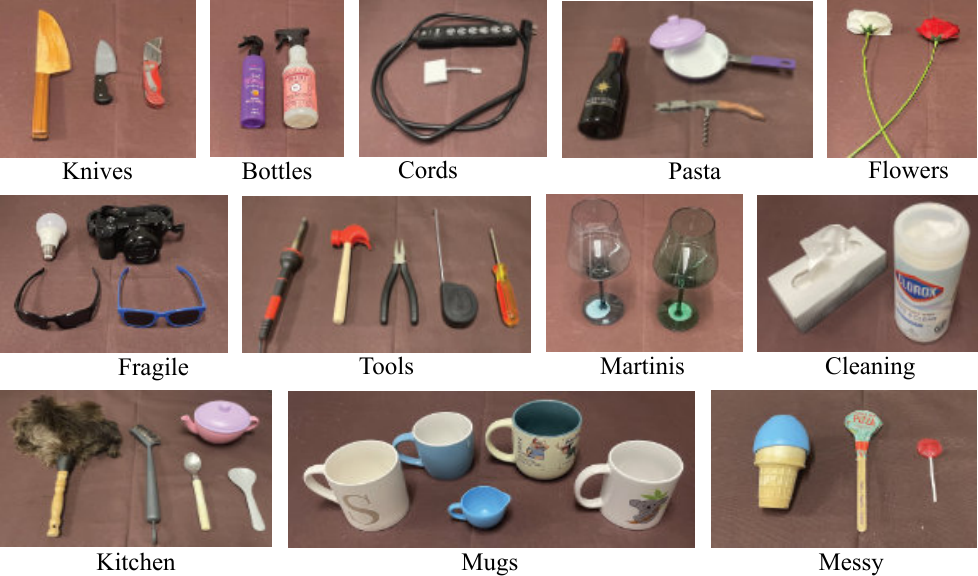}
    \caption{Overview of the $38$ objects used in our study, inspired by the objects introduced in LERF-TOGO~\cite{lerftogo2023}.}
    \label{fig:eobjects}
\end{figure}

To select an appropriate grasping point for the object, we propose to leverage the commonsense knowledge encoded in LLMs across two interaction stages, inspired by the Chain of Thought approach~\cite{wei2022chain}, further improving our approach's reasoning capabilities: 
$1)$ semantic reasoning over each shape described in the graph $\gG$ and 
$2)$ selecting the most appropriate shape that facilitates the task-oriented grasp. 
We leverage a prompt template, described in Fig.~\ref{tab:prompt}, that depends only on the target object, desired task, and constructed graph $\gG$. To ensure the intended output structure, we utilize TypeChat~\cite{TypeChat} in addition to the presented prompts.

\subsubsection{Semantic Part Identification}
In the first stage, given the target object name, task, and constructed graph, the LLM is instructed to reason about the nodes in the graph and what semantic part (e.g., ``handle'' and ``blade'', for a knife) each represents in the target object.
To conduct this reasoning, we first ask for an unstructured, free-form answer in which the LLM explicitly explains its thoughts.
As a follow-up to this response, the LLM is tasked to assign a single semantic part to each graph node in a structured manner.

\subsubsection{Task-Oriented Part Selection}
After the semantic reasoning is complete, the LLM is instructed to reason about the task utility of each part, given its graph representation and assigned semantics from the first stage. Similar to the first stage, this is accomplished in two steps: a free-form reasoning and explanation stage, which the LLM then uses to assign a final task-oriented suitability score to each node, which determines the selected grasp.

\subsection{Selecting a Grasp Pose}
\label{sec:grasp_pose}
In the final step, we select the graph node with the highest predicted score. 
To facilitate the grasp, we calculate the centroid of the chosen part and derive its corresponding 3D coordinates 
from the depth information in input image $\mI$.
To consider rotations, we calculate the principle components of the masked subpart in the point cloud and grasp along the largest component.



\section{Experiments}
\label{sec:experiments}

We evaluate our approach in real-world experiments and demonstrate its effectiveness in grasping a diverse range of household objects across a variety of tasks.
Figure~\ref{fig:eobjects} presents the $38$ objects from $12$ different categories used in the $49$ tasks in our experiments. 
Section~\ref{sec:decomposition} discusses our heuristic that both dynamically sets the 2D and 3D decomposition's threshold (Sec.~\ref{sec:threshold}) and selects between their final outputs (Sec.~\ref{sec:2dv3d}).
Section~\ref{sec:results} demonstrates our approach on a real-world robotic platform and compares it against state-of-the-art baselines, and also includes additional qualitative experiments in section~\ref{sec:qual_results} and an LLM ablation study in section~\ref{sec:ablation}.
All of our experiments are conducted with a Kinova Jaco robotic arm equipped with a three-finger gripper, coupled with a fixed Oak-D SR passive stereo-depth camera for RGB and depth perception. 

\textbf{Pipeline Configuration:} To retrieve convex decompositions of sufficient quality, we empirically set the threshold $\omega$, deciding when to select the 2D decomposition over the 3D decomposition, to $\omega = 10$. Additionally, we set parameter $\alpha$, defining the minimal percentage of depth points that exhibit high confidence, to $\alpha = 85\%$.
We note these values are not sensitive and are set emprically without intricate tuning.

\textbf{Dataset:} We created a dataset of $38$ objects covering $12$ general categories and $49$ tasks, as shown in Fig.~\ref{fig:eobjects}, inspired by the objects and tasks used in the LERF-TOGO~\cite{lerftogo2023} dataset.

\textbf{Metrics:}  To evaluate the effectiveness of our approach, we employ three metrics: ``Part Identification'', ``Part Selection'', and ``Lift Success''. 
``Part Identification'' measures the accuracy of the semantic part assignment, across all parts globally and across only the ground truth parts. ``Part Selection'' quantifies the proficiency of our model to select the correct part for the task-oriented grasp. 
For this metric, the selection is deemed correct if it aligns with ground-truth parts based on LERF-TOGO~\cite{lerftogo2023}'s part queries and commonsense human judgment with respect to the task, safety, and stability.
``Lift Success'' indicates the percentage of successfully lifted objects at the selected part for the given task. Incorrectly selected parts and failed grasps at correctly selected parts are both counted as failures here.




\subsection{Automatic Geometric Decomposition}
\label{sec:decomposition}




\subsubsection{Dynamic Threshold Selection}
\label{sec:threshold}

\begin{figure}
    \centering
    \includegraphics[width=1\linewidth]{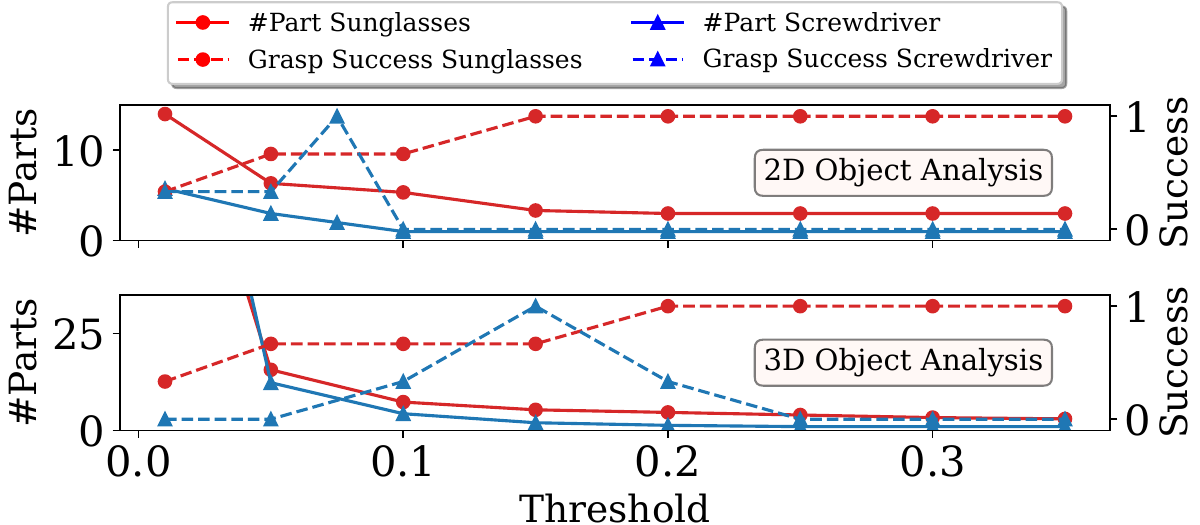}
    \caption{Threshold analysis for 2D (top)/3D (bottom) decompositions on sunglasses (complex) and screwdriver (simple).}
    \label{fig:combined-analysis}
\end{figure}

In both the 2D and 3D pipelines, the convex decomposition threshold $\gamma$ is an important element that facilitates grasping success by controlling the number of decomposed object parts. This threshold plays a pivotal role in balancing the accuracy of the decomposition with the manageability of the resulting segmentation. A high error threshold $\gamma > 0.2$ that does not over-segment a geometrically complex object may fail to decompose a simple object -- the entire object may be approximated as a single convex hull. Conversely, a low error threshold $\gamma < 0.1$ may decompose complex objects into overly many parts, complicating the resulting graph and making reasoning challenging. As shown in Figure~\ref{fig:combined-analysis}, we run the pipeline at thresholds between $0.01$ and $0.35$ and evaluate the number of decomposed parts and the lifting success rate on sunglasses and screwdriver, respectively representing geometrically complex and simple objects, as evidenced by the consistent larger number of decomposed parts for the sunglasses at each threshold. This experiment reveals that any single threshold may not be sufficient to accommodate all objects. At thresholds conducive to successful reasoning over and grasping the sunglasses, the screwdriver often fails to properly decompose; naively taking the centroid of an entire object as the grasp point does not allow room for reasoning and is inherently unsafe, which we count as a failure. While this threshold may easily be set and tuned by a user, our heuristic fully automates the inference process.

Our pipeline provides a generalizable zero-shot approach for grasping by balancing object complexities and geometries. Therefore, we propose the selection algorithm described in Section~\ref{sec:extracting_shapes} that considers the number of decomposed parts to set an object-specific threshold $\gamma$. An initial threshold is set at $\gamma = 0.2$ for 3D meshes and $\gamma = 0.15$ for 2D meshes. The error allowance for 3D objects should naturally be higher to accommodate potential noise in the depth perception. These thresholds are set conservatively and tend to under-decompose objects. To address this, we iteratively decrease the threshold by $0.025$ for each object until a valid decomposition is achieved (an object is decomposed into two or more parts).
For example, a valid decomposition and successful lift are achieved within $2-3$ iterations for the screwdriver, while the initial threshold is satisfactory for sunglasses. 
We further find that in our experiments, high thresholds that do not decompose an object execute in less than a second, making this search both efficient and effective. 

\subsubsection{2D vs 3D Decompositions}
\label{sec:2dv3d}

After setting appropriate thresholds, the choice between the resulting 2D and 3D decompositions used to build the graph is a critical step. As discussed in Section~\ref{sec:extracting_shapes}, the heuristic algorithm we employ uses the depth confidence and number of decomposed parts (too many parts may indicate noisy, concave, or overly complex surfaces) to facilitate this choice. Table~\ref{tab:variants} shows the ``Part Selection'' performance across all object-task pairs between the 2D and 3D pipelines and heuristic that selects between them, each of which uses the heuristic threshold search. An interesting observation is that while our heuristic selection does indeed result in the best performance ($92\%$), the 2D pipeline exhibits stronger performance ($86\%$) than the 3D pipeline ($83\%$). While this may be initially counterintuitive, this result demonstrates and is consistent with the conclusion that the flexibility of our pipeline allows us to dynamically adapt to settings where depth information may be inconsistent, low quality, or incompatible with the convex decomposition method. In our real-world experiments, while the Oak-D SR passive depth camera provides high-quality depth estimates in well-lit environments on matte objects with noticeable elevated features---the same features we may care to segment for grasping---results deteriorate with concave, reflective, and transparent surfaces, where the 2D pipeline excels due to the sole reliance on RGB data. The complexity and prevalent noise in real-world settings often necessitates a reliance on 2D decompositions for accuracy and robustness while still leveraging depth data in 3D decompositions when it is confident and useful via the heuristic.

The threshold search, coupled with the automatic selection between the 2D and 3D pipelines, forms an algorithm that determines decomposition outputs that are suitable and useful for semantic reasoning and grasping.

\subsection{Zero-Shot Task-Oriented Grasping}
\label{sec:results}

For real-world task-oriented grasping experiments, we pair the \texttt{ShapeGrasp} pipeline with a Kinova Jaco arm designed to grasp and lift objects at the selected part (see Table~\ref{tab:part_results}). This setup employs a straightforward method for selecting grasp poses drawing directly from the object graph and masked point cloud (see ~\ref{sec:grasp_selection}) to execute a top-down grasp.

We employ two baselines to evaluate the efficacy of our proposed approach \texttt{ShapeGrasp} on the ``Part Selection'' and ``Lift Success'' metrics:
\textbf{GraspGPT \cite{tang2023graspgpt}}, which is a current state-of-the-art approach for zero-shot task-oriented grasping and 
\textbf{GPT4-V \cite{Achiam2023GPT4TR}}, a foundation model trained with internet-scale data with visual input modality, prompted with language instructions to select the correct task-oriented part.

\begin{table}[]
    \centering
    \begin{tabular}{@{\makebox[1em][r]{\scriptsize\rownumber\space}} lcc}
        \toprule
        \multicolumn{1}{c}{Model} \textit{(using GPT-4)}        & Part Selection \\
        \midrule
        \texttt{ShapeGrasp} (2D only)   & $0.86$  \\
        \texttt{ShapeGrasp} (3D only)   & $0.73$  \\
        \midrule
        \texttt{ShapeGrasp} (Heuristic, no obj)   & $0.51$  \\
        \texttt{ShapeGrasp} (Heuristic)   & $\mathbf{0.92}$  \\
        \bottomrule
    \end{tabular}
    \caption{Part selection accuracy on different variants of \texttt{ShapeGrasp}. We evaluate 2D-only, 3D-only, and heuristic decompositions. For the heuristic, we also explore a ``no obj'' variant that omits the object name from the prompt and task.}
    \label{tab:variants}
\end{table}

\begin{table}[]
    \setlength{\tabcolsep}{1.5mm}
    \centering
    \begin{tabular}{@{\makebox[1em][r]{\scriptsize\rownumber\space}} lcccc}
        \toprule
        \multicolumn{1}{l}{Model}   & Part ID    & Part Sel. & Success & Time \\
        \midrule
        GraspGPT \cite{tang2023graspgpt}  & N/A & $0.37$ & $0.31$ & $150$\\
        GPT4-Vision \cite{Achiam2023GPT4TR}    & N/A & $0.82$ & $0.73$ & $20$\\
        \midrule
        \texttt{ShapeGrasp} (Starling)  & $0.54$ $(0.63)$& $0.65$ & $0.57$ & $25$\\
        \texttt{ShapeGrasp} (GPT-4)  & $\mathbf{0.84}$ $\mathbf{(0.90)}$ & $\mathbf{0.92}$ & $\mathbf{0.82}$ & $30$\\
        \bottomrule
    \end{tabular}
    \caption{Results on \texttt{ShapeGrasp} compared to GraspGPT and GPT4-V baselines. Part ID is the semantic ``Part Identification'' accuracy across all parts (and across the target part), Part Sel. is the ``Part Selection'' accuracy, ``Success'' is the robot's ``Lift Success'', and ``Time'' is the typical inference time in seconds for each method.}
    \label{tab:part_results}
\end{table}

Our empirical findings indicate a significant performance advantage of our method over GraspGPT \cite{tang2023graspgpt}, underscoring the efficacy of our structured, symbolic object part graph in conjunction with LLM reasoning. The performance gain using our pipeline over GraspGPT \cite{tang2023graspgpt} is $55\%$ and $51\%$ (see rows $1$ and $4$ in Table~\ref{tab:part_results}) for the ``Part Selection'' and ``Lift Success'' metrics, respectively. 
We further analyze the discrepancy between \texttt{ShapeGrasp} and GraspGPT results; we note that GraspGPT 
depends on GraspNet~\cite{mousavian20196} for grasp sampling, which may be inaccurate or fail on certain objects when the depth quality is noisy or poor, which may occur due to our static monocular depth camera. 
While GraspGPT is limited to tasks and objects related to previously known concepts, 
\texttt{ShapeGrasp} demonstrates robustness to the same noisy depth inputs, while featuring zero-shot and being more lightweight than GraspGPT (see ``Time'' in Table~\ref{tab:part_results}).

An important hypothesis that motivates our vision-based pipeline that constructs the object graph is that directly processing object part features and spatial relationships, and providing this information in a structured way for LLM reasoning, is more robust and performant than relying on VLMs for end-to-end reasoning. Though VLMs are considerably larger and more expensive models, performance on low-level features and relationships within parts of an object image may be unreliable and subject to hallucinations \cite{yuksekgonul2023when}.

To test this hypothesis and directly compare our graph-construction and reasoning pipeline to a VLM, we establish a privileged GPT4-Vision baseline that benefits from the same heuristic-selected object segmentations and skips the graph-based reasoning stages (contributing to the faster runtime). This baseline is grounded by coloring each part and assigning them integer index labels for clarity. We confirm GPT4-Vision's capability to interpret segmented and grounded input object images through a series of questions and human-verified responses. We use the same method to determine the grasp pose for the GPT4-Vision selected part to ensure comparability. Our method shows significant success rate gains over GPT4-Vision, by $10\%$ and $9\%$ on the evaluation metrics (see rows $2$ and $4$ in Table~\ref{tab:part_results}).

We further test the modularity of \texttt{ShapeGrasp} by evaluating the full pipeline using Starling~\cite{starling2023}, a much smaller and more efficient open-source LLM, as the inference backend instead of GPT-4. Performance across all metrics, while lower than the larger and more powerful GPT-4, remain meaningful and higher than the GraspGPT \cite{tang2023graspgpt} baseline.

\subsubsection{Qualitative Results}
\label{sec:qual_results}
\begin{figure}
    \centering
    \includegraphics[width=1\linewidth]{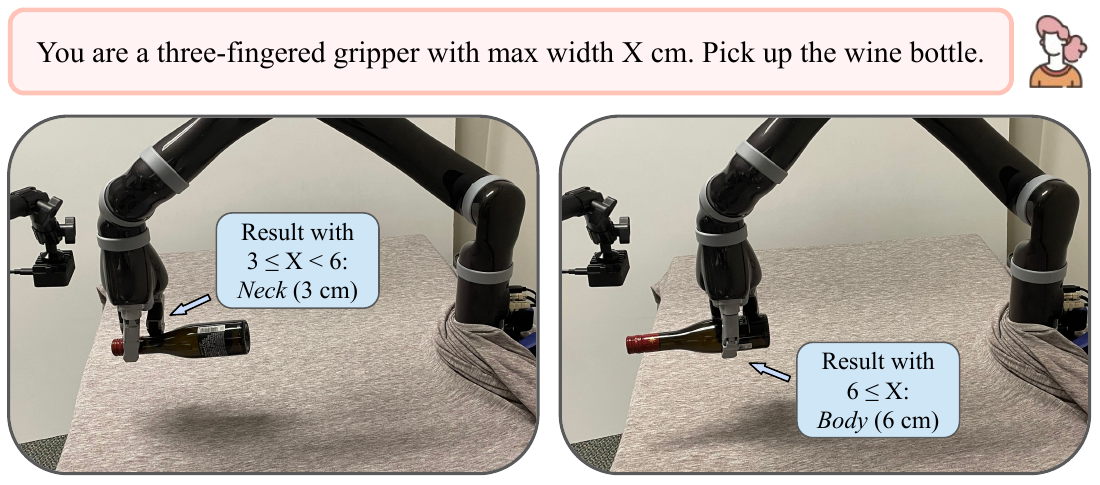}
    \caption{\texttt{ShapeGrasp} results incorporating width attributes and variable gripper width constraints.}
    \label{fig:gripperwidth}
\end{figure}

The flexibility and generalizability of \texttt{ShapeGrasp} allow us to explore more complex interactions by incorporating additional object and robot attributes.
For example, Fig.~\ref{fig:tog} shows how LLM semantic reasoning over our shape graph enables effective execution of the ``hand over'' task. As the LLM understands which node in the graph corresponds to the handles in objects (semantic identification stage), it can prioritize that part in the ``hand over'' interaction (task-oriented selection stage). Additional attributes can also be incorporated and reasoned over in an object-specific way; while both the ``mug'' and the ``soldering iron'' are given the attribute of being ``hot'', the task-oriented reasoning stage can make the commonsense inference that the level of heat and risk exhibited by these two objects differ dramatically. For the ``hand over'' task, while the mug is grasped by the hot body, which ``minimizes the risk of spilling hot liquid and ensures a comfortable handover'', the soldering iron is still grasped by the handle, which ``positions the hot tip away from both the robot and the human'' (see Fig.~\ref{fig:tog}).  

Fig.~\ref{fig:gripperwidth} demonstrates how the task-oriented selection stage is also able to consider robot attributes. Here, a ``width'' attribute is added to each node in the object graph, along with a variable maximum gripper width in the prompt. While the robot prefers to ``pick up the wine bottle'' by the bottle body, when the body width exceeds the maximum gripper width, it switches to the narrower neck, ensuring a successful grasp.

\subsubsection{LLM Ablations}

\label{sec:ablation}
To validate the efficacy of our model and the comprehensiveness of the LLM interaction stages, we conducted an ablation study with a focus on the ``Part Selection'' evaluation metric, as detailed in Table~\ref{tab:ablation_study}. Applying the previously described heuristic decomposition selection algorithm, we examine the impact of each LLM reasoning stage: the \textit{semantic part identification} stage and \textit{task-oriented part reasoning} stage that comes before the final task score assignments. We evaluate the results using neither reasoning stage (LLM directly assigns only task scores), only one reasoning stage, and the full reasoning procedure.

The ablation study indicates that the most effective performance is achieved when both the part identification and task reasoning stages are employed in the LLM interaction ($92\%$). The part identification stage is empirically the more important of the two stages, leading to a performance of ($87\%$). This stage is likely important because when employed, the LLM's final selection is forced to depend on its own semantic part assignments. Intuitively, the semantics of a part are essential in determining its affordances and suitability for a task. This argument is further supported by the accuracy in the part identification stage (see Table~\ref{tab:part_results}), $84\%$ globally, and $90\%$ across the ground truth parts. These numbers lead to an important observation: not every part in an object, or even the ground-truth part, needs to be correctly identified; it may be sufficient if even a single critical part to either grasp or avoid is correctly identified. For example, in a pair of blue sunglasses, the ground truth part ``arm'' was misidentified as the ``frame'', but the ``lens''---which is critical to avoid---was correctly identified. As such, the task-oriented selection stage correctly prioritized avoiding the ``lens'' segment. Put in another way, the quality of the automatic heuristic decomposition and the accuracy of the part identification stage effectively provides another novelty: \textit{semantic part segmentation}, which naturally may be correlated with task-oriented part selection. Note that the ``Part Identification'' and ``Part Selection'' accuracies are closely coupled with both the GPT-4 and Starling models.

Including the task-oriented reasoning stage demonstrates a notable performance gain as well, both when the part-identification stage is ablated ($14\%$) and included ($8\%$); this may indicate that even once semantics are assigned, explicit task-conditioned reasoning is still beneficial, especially as a single object may have numerous appropriate uses and tasks.

In Table~\ref{tab:variants}, we also explore the performance when the target object name is withheld. This leads to a significant performance drop ($51\%$), likely due to the challenge of assigning semantic parts to an unknown object and highlighting \texttt{ShapeGrasp}'s effective use of this basic information. In this setting, assigned semantics may not be meaningful, with the results largely determined by geometric attributes.

\begin{table}[]
    \centering
    \begin{tabular}{@{\makebox[1em][r]{\scriptsize\rownumber\space}} ccc}
        \toprule
        \multicolumn{1}{c}{Task Reasoning} & Part Identification & Part Selection \\
        \midrule
        \centering &   & $0.43$ \\
        \midrule
        \centering\cmark &    & $0.57$ \\
        \centering & \cmark   & $0.84$ \\
        \midrule
        \centering\cmark & \cmark   & $\mathbf{0.92}$ \\
        \bottomrule
    \end{tabular}
    \caption{Ablation study on the LLM reasoning within \texttt{ShapeGrasp}. While a part to grasp is always produced, seasoning can be done without identifying parts and/or without reasoning over the task. See Fig.~\ref{tab:prompt} for detailed prompt information. }
    \label{tab:ablation_study}
\end{table}

\section{Conclusion, Limitations, and Future Work} 
In this work, we present \texttt{ShapeGrasp}, a novel approach that performs fine-grained semantic reasoning over an object's geometric composition to find a suitable task-oriented part and associated grasp.
Our novel representation of an object's geometric composition as a graph of basic shapes, allows an LLM to effectively reason over the semantic significance and task utility of each part to ultimately select the most suitable part for the desired task.
Through extensive experiments on real-world hardware, we demonstrated that our approach can efficiently utilize a single, static RGB+D camera image for zero-shot task-oriented grasping and outperform current state-of-the-art approaches. 

We recognize that the synergy of convex part decomposition and the LLM's ability to assign semantic meaning to each segment is a powerful feature of our method while, at the same time, introducing a dependency on a ``reasonable'' decomposition. 
While we mitigate this dependency through the introduction of our heuristic that automatically tunes the decomposition pipeline, further integration into alternative decomposition methods may further improve our pipeline.
In future work, we will investigate the utility of our approach with even less information, particularly when no semantic object information is provided (initial results in Table~\ref{tab:variants}, row $3$), as the geometric attributes of an object's decomposed parts alone may be a powerful reasoning signal.
Such improved reasoning could be facilitated by elevating our graph construction into the third dimension and utilizing 3D-geometries.
Further, we will investigate different LLM prompting approaches to improve reasoning capabilities.

\section*{Acknowledgements}
We want to acknowledge the support from DARPA under grant FA8750-23-2-1015, AFOSR under grants FA9550-18-1-0251 and FA9550-18-1-0097, and ARL under grant W911NF-19-2-0146 and W911NF-2320007.

\bibliographystyle{IEEEtran}
\bibliography{main}

\end{document}